\documentclass[letterpaper, 10 pt, conference]{ieeeconf}  

\IEEEoverridecommandlockouts                              

\usepackage{xspace}
\usepackage{cite}
\usepackage{amsmath,amssymb,amsfonts}
\usepackage{bm}
\usepackage{booktabs}
\usepackage[inkscapearea=drawing, inkscapepath=figures/generated, inkscapelatex=false]{svg}
\usepackage[caption=false, font=footnotesize]{subfig}
\usepackage{algorithm}
\usepackage{algorithmic}
\usepackage{graphicx}
\usepackage{textcomp}
\usepackage{xcolor}
\usepackage{tikz}
\usepackage{eso-pic}

\let\labelindent\relax 
\usepackage{enumitem}
\usepackage[hidelinks]{hyperref}
\usepackage[symbol,hang,flushmargin]{footmisc}
\usepackage[normalem]{ulem}
\usepackage[nolist]{acronym}
\usepackage{listings}
\usepackage{float}
\usepackage{comment}
\usepackage{subfig}
\acrodef{spi}[SPI]{Shadow Program Inversion}
\acrodef{dof}[DoF]{degree of freedom}
\acrodefplural{dof}[DoF]{degrees of freedom}
\acrodef{rl}[RL]{Reinforcement Learning}
\acrodef{tcp}[TCP]{tool center point}
\acrodef{dmp}[DMP]{Dynamic Movement Primitive}
\acrodef{prodmp}[ProDMP]{Probabilistic Dynamic Movement Primitive}
\acrodef{iql}[IQL]{Implicit Q-Learning}
\acrodef{awr}[AWR]{Advantage Weighted Regression}
\acrodef{bc}[BC]{Behaviour Cloning}
\acrodef{dcg}[DCG]{differentiable computation graph}
\acrodef{vit}[ViT]{Vision Transformer}
\acrodef{xai}[XAI]{explainable artificial intelligence}
\acrodef{dp}[$\partial$P]{differentiable programming}
\acrodef{ann}[ANN]{artificial neural network}
\acrodef{llm}[LLM]{Large Language Model}
\acrodef{cad}[CAD]{Computer-Aided Design}
\acrodef{cot}[CoT]{Chain-of-Thought}
\acrodef{gnn}[GNN]{Graph Neural Network}
\acrodef{sam}[SAM]{Segment Anything Model}
\acrodef{api}[API]{Application Programming Interface}

\newcommand{\querycad}{QueryCAD\xspace}
\newcommand{\segcad}{SegCAD\xspace}
\newcommand{\cadds}{CAD-Q\&A\xspace}
\newcommand{\linktowebsite}{\url{https://claudius-kienle.github.io/querycad}}

\definecolor{lbcolor}{rgb}{0.95,0.95,0.95}  
\title{\LARGE \bf \querycad: Grounded Question Answering for CAD Models}

\author{Claudius Kienle$^{1,2}$, Benjamin Alt$^{1,3}$, Darko Katic$^{1,4}$, Rainer Jäkel$^{1}$ and Jan Peters$^{2}$
\thanks{This  work  was  supported  by  the  German  Federal  Ministry  of  Education and Research (grant 01MJ22003B), the KMU-innovative project RobInTime and the Learning Factory Global Production at the wbk Institute of Production Science at Karlsruhe Institute of Technology.}
\thanks{$^{1}$ArtiMinds Robotics, Karlsruhe, Germany}%
\thanks{$^{2}$IAS Lab, Computer Science Department, TU Darmstadt, Germany}%
\thanks{$^{3}$AICOR Institute for Artificial Intelligence, University of Bremen, Germany}%
\thanks{$^{4}$Stuttgart University of Applied Sciences, Germany}%
}

\newcommand\copyrighttext{%
  \footnotesize \textcopyright 2025 IEEE. Personal use of this material is permitted.  Permission from IEEE must be obtained for all other uses, in any current or future media, including reprinting/republishing this material for advertising or promotional purposes, creating new collective works, for resale or redistribution to servers or lists, or reuse of any copyrighted component of this work in other works.}

\newcommand\copyrightnotice{%
  \AddToShipoutPictureBG*{%
    \AtPageLowerLeft{%
      \raisebox{22pt}[0pt][0pt]{%
        \makebox[\paperwidth]{%
          \hfill\fbox{%
            \parbox{\dimexpr\textwidth-\fboxsep-\fboxrule\relax}{\copyrighttext}%
          }\hfill
        }%
      }%
    }%
  }
}

\begin{document}
\bstctlcite{IEEEexample:BSTcontrol}

\maketitle
\copyrightnotice
\thispagestyle{empty}
\pagestyle{empty}

\begin{abstract}
    CAD models are widely used in industry and are essential for robotic automation processes. 
    However, these models are rarely considered in novel AI-based approaches, such as the automatic synthesis of robot programs, as there are no readily available methods that would allow CAD models to be incorporated for the analysis, interpretation, or extraction of information. 
    To address these limitations, we propose \querycad, the first system designed for CAD question answering, enabling the extraction of precise information from CAD models using natural language queries. \querycad incorporates \segcad, an open-vocabulary instance segmentation model we developed to identify and select specific parts of the CAD model based on part descriptions. We further propose a CAD question answering benchmark to evaluate \querycad and establish a foundation for future research. Lastly, we integrate \querycad within an automatic robot program synthesis framework, validating its ability to enhance deep-learning solutions for robotics by enabling them to process CAD models. \linktowebsite %
\end{abstract}

\section{Introduction}
\label{sec:introduction}

In the industrial sector, many workflows are centered around \ac{cad} models. These models contain precise representations of individual parts and their assembly into larger components. Engineers in various industrial domains, such as robot programming, rely heavily on \ac{cad} models to extract measurements, identify specific features, and understand how parts interact within a larger system. To automate these still largely manual engineer\-ing processes, it is crucial to develop methods that can retrieve and interpret information from \ac{cad} models in an automated manner, mirroring the way users interact with them. 
However, since no web-scale dataset for industrial \ac{cad} models exists, it is generally hard to train or extend existing models to understand this modality \cite{colligan_hierarchical_2022}. Moreover, there currently exists no method for automatic information retrieval from \ac{cad} models. This gap is particularly evident in the context of automated robot program synthesis for industrial applications, where extracting parameters from \ac{cad} models is essential and done manually.

\begin{figure}
    \centering
    \includesvg[width=\linewidth]{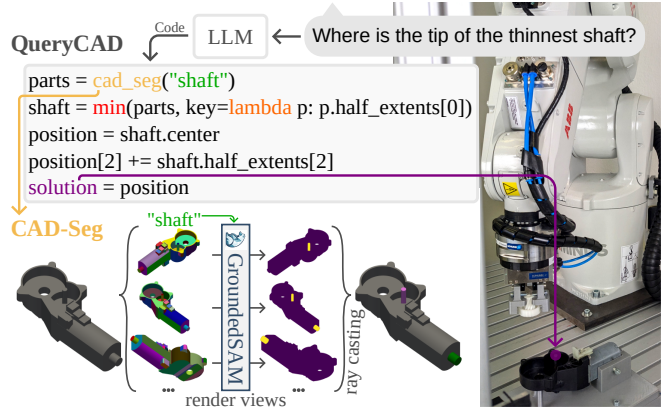}
    \caption{
    \querycad computes precise measurements in response to natural-language queries. It leverages a code-writing \ac{llm} with in-context prompting to generate executable code, which then retrieves specific parts of the \ac{cad} model through multi-view segmentation. This process allows for extraction of detailed measurements from \ac{cad} models based on the high-level user queries. The computed measurements can be utilized in downstream applications such as automatic robot program parameterization, CNC machining or 3D printing.}
    \label{fig:page-1}
\end{figure}

In response to this, we propose \querycad, the first deep learning-based question-answering system for \ac{cad} models. \querycad is the first system to retrieve specific information like measurements, positions, or counts of features or parts of \ac{cad} models in response to natural-language questions. Given a free-text question, \querycad generates answers by grounding its reasoning process in structured engineering knowledge, directly retrieving measurements from the \ac{cad} model (see Fig. \ref{fig:page-1}). Grounding natural-language queries in \ac{cad} data enables seamless integration of \ac{cad} models into deep learning-based applications and is particularly compatible with existing frameworks that build upon \acp{llm} \cite{liang_code_2023,alt_robogrind_2024}. \querycad 
 is integrated into the MetaWizard robot program synthesis system \cite{alt_robogrind_2024} to enable the automatic generation of industrial robot programs given natural-language task specifications.

In this paper, we make the following contributions:
\begin{enumerate}
    \item \textbf{\segcad}: A model for open-vocabulary multi-view \ac{cad} part segmentation.
    \item \textbf{\querycad}: The first system for \ac{cad} question answering.
    \item \textbf{\cadds Benchmark}: A benchmark for \ac{cad} question answering, containing 111 questions about 18 different \ac{cad} models.
    \item \textbf{Robot Program Synthesis}: Integration of \querycad in a robot program synthesis framework for automatic generation of robot programs grounded in \ac{cad} data.
\end{enumerate}

\section{Related Work}

%

\subsection{3D Part Segmentation}
3D part segmentation focuses on methods for partitioning 3D objects represented as voxels, meshes, or point clouds into their functional components, e.g. separating a hammer into its handle and head. These methods can be categorized based on their input modalities, whether they utilize prompts for segmentation, as well as the type of prompts used.

Qi et al. \cite{qi_pointnet_2017,qi_pointnet_2017-1} introduced PointNet, a neural network designed to segment point clouds into coherent parts and trained on extensive part segmentation datasets. Subsequent work by Qian et al. \cite{qian_pointnext_2022} enhance PointNet’s performance by optimizing its training procedure and scaling strategies.

To specify the parts to be segmented, some methods leverage text-based prompts to guide part segmentation. Liu et al. \cite{liu_partslip_2023,zhou_partslip_2023} utilize the pretrained image-language model GLIP \cite{li_grounded_2022}, which benefits from its extensive training on web-scale images. They employ a multi-view detection strategy to convert 2D bounding boxes into 3D segmentations, enhancing zero-shot generalization and open-set detection. The language modality of GLIP enables specification of segmented parts using text-based descriptions.

Another recent advancement involves using point-based prompts for segmentation. Zhou et al. \cite{zhou_point-sam_2024} introduce a 3D segmentation model capable of segmenting parts based on a point provided as prompt. This approach allows to specify the part to segment by indicating a point on the object. 

There is a wide range of methods for part segmentation, each utilizing different representations and techniques. Many approaches focus on point clouds \cite{qi_pointnet_2017,qi_pointnet_2017-1,liu_partslip_2023,zhou_partslip_2023,zhou_point-sam_2024}, while some also leverage voxel-based representations \cite{graham_3d_2018}. However, no current methods directly segment \ac{cad} models into parts. Furthermore, existing approaches do not segment parts based on text-based descriptions. Recent methods predict bounding boxes \cite{liu_partslip_2023,zhou_partslip_2023}, which is too imprecise for \ac{cad} models where more detailed retrieval via segmentation masks is needed.

\subsection{3D Scene Understanding}
In research, there is growing interest in solving tasks related to 3D scenes. Most prominently, 3D question answering revolves around answering natural-language questions about objects in a scene \cite{huang_chat-3d_2023,zhu_3d-vista_2023,parelli_clip-guided_2023,azuma_scanqa_2022,chen_scanrefer_2020,ma_sqa3d_2023}. Huang et al. \cite{huang_chat-3d_2023} propose an \ac{llm}-based model that works with object identifiers of a 3D scene and can answer questions about objects as well as identify the object asked about in the question. 
3D visual grounding \cite{huang_chat-3d_2023,wang_3drp-net_2023,zhao_3dvg-transformer_2021,wang_distilling_2023,chen_language_2022,huang_multi-view_2022,chen_scanrefer_2020} is the task to locate a object in a 3D scene by natural language. Zhao et al. \cite{wang_3drp-net_2023} propose a model that processes the raw point cloud and fuses it with the text to predict a bounding box that surrounds the target object. The finetuned \ac{llm} proposed by Huang et al. \cite{huang_chat-3d_2023} was also applied for visual grounding by determining the object identifier that matches with the natural language description of the target object.

There is a considerable body of work on 3D (indoor) scene question answering and integrating natural language with point-cloud represented scenes. However, to the best of our knowledge, no work exists on transferring these approaches for question answering tasks on \ac{cad} models or parts.


\subsection{CAD Machining Feature Classification}
Numerous approaches have been developed for the automatic classification of operations required for machining a \ac{cad} model \cite{cao_graph_2020,cha_machining_2023,lambourne_brepnet_2021,jayaraman_uv-net_2021,feng_meshnet_2019,hanocka_meshcnn_2019,yang_dsg-net_2022,colligan_hierarchical_2022,lei_mfpointnet_2022,zhang_featurenet_2018,wu_3d_2015,peddireddy_deep_2020}. These methods aim to assign specific machining features to each face of the \ac{cad} model from a predefined set of features. The approaches can be categorized based on the modality the classification model operates on.

One common strategy is to represent the \ac{cad} model as a graph, with two prevalent types of graph representations found in the literature. In a \textit{BRep-Graph}, faces are represented as nodes, where two nodes are interconnected if they are adjacent on the \ac{cad} model \cite{cao_graph_2020,cha_machining_2023,lambourne_brepnet_2021,jayaraman_uv-net_2021}.
A \textit{Facet-Graph} represents the facets of the triangulated mesh of the \ac{cad} model \cite{feng_meshnet_2019,hanocka_meshcnn_2019,yang_dsg-net_2022,colligan_hierarchical_2022}. The node and edge features vary across different approaches but consistently include properties of the face or facet. To classify the nodes, these approaches employ \acp{gnn}. Notably, Colligan et al. \cite{colligan_hierarchical_2022} combine a \textit{BRep-Graph} and a \textit{Facet-Graph} into a hierarchical \ac{gnn} to classify each node according to its machining feature.

In another set of approaches, the \ac{cad} model is sampled to a point cloud \cite{lei_mfpointnet_2022} or voxel grid \cite{zhang_featurenet_2018,wu_3d_2015,peddireddy_deep_2020}. These methods then train a 3D segmentation network to classify each point or voxel by its corresponding machining feature. For example, Lei et al. \cite{lei_mfpointnet_2022} sample a point cloud from the \ac{cad} model and train a 3D segmentation network to classify each point into one of 33 machining features.

All approaches share the common requirement of training a neural network for the classification task. This necessitates a large dataset, which is often generated synthetically \cite{colligan_hierarchical_2022,zhang_featurenet_2018,cao_graph_2020} because existing large-scale real-world datasets \cite{koch_abc_2019,kim_large-scale_2020} lack the necessary labels \cite{colligan_hierarchical_2022}. While these methods are highly effective for machining classification and show a good generalization to new shapes, they are limited by their closed-set classification, making it difficult to adapt them for identifying additional feature types. Furthermore, these approaches typically operate at a low feature level, focusing on individual machining features rather than classifying higher-level features like parts or geometric structures.


\section{Methods}



\begin{figure*}
    \centering
    \includesvg[width=\linewidth]{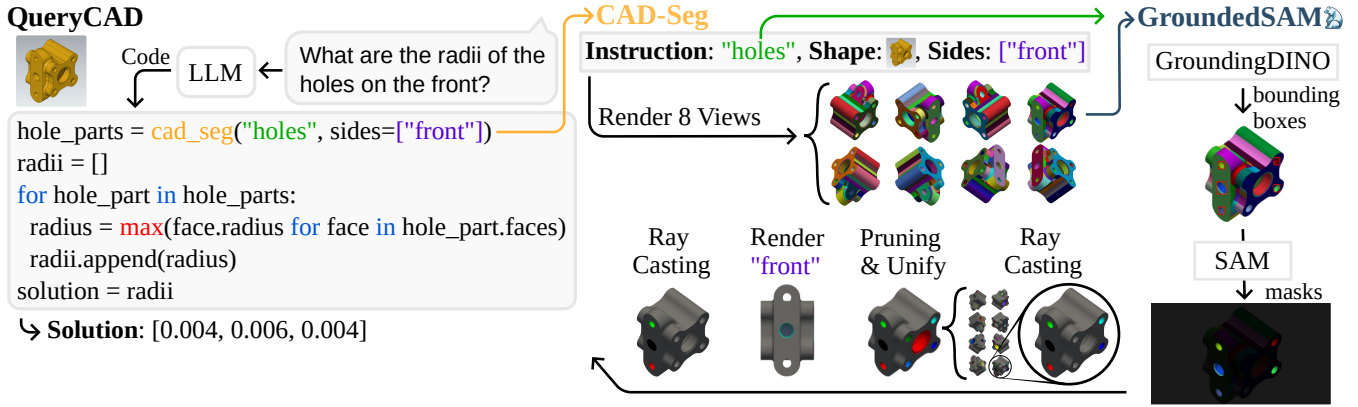}
    \caption{\querycad: Given a free-form question, the \ac{llm} generates Python code to compute the required measurements or properties of the \ac{cad} model. The generated code invokes \segcad to identify the relevant parts of the \ac{cad} model specified in the question. \segcad renders the \ac{cad} model from multiple views and performs ray casting to align the rendered images with the \ac{cad} faces. Parts that match the description, such as ``holes'', are retrieved and returned as Python objects for further processing. The final answer is computed by the generated Python code and stored as variable \textit{solution}.}
    \label{fig:architecture}
\end{figure*}

We propose \querycad, a system to answer free-form questions about \ac{cad} models and their parts. To acomplish this, \querycad utilizes \segcad, a model for segmenting parts of a \ac{cad} model matching an open-vocabulary part description. The architecture can be seen in Figure \ref{fig:architecture}. To evaluate \querycad, we develop the first benchmark for the task of \ac{cad} question answering.

\subsection{\segcad: Open-Set CAD Segmentation}

In \ac{cad} question answering, the goal is to retrieve specific information about parts or individual features of a \ac{cad} model. This process begins with identifying the parts or features referenced in a given question. To achieve this, we developed \segcad, a \ac{cad} instance segmentation model specifically designed to identify parts or features based on free-text part descriptions.

The \ac{cad} model is first \textit{3D rendered} (\ref{sec:segcad-3drendering}) from predefined viewing angles, with each rendered 2D image showing the \ac{cad} model from a unique perspective. These images are then processed through a promptable \textit{instance segmentation} model \cite{ren_grounded_2024} (\ref{sec:segcad-inst-seg}), which masks the parts in the image that correspond to the provided part description. Finally, the masked image is unprojected back into 3D space via \textit{image to \ac{cad} model alignment} (\ref{sec:img-to-cad-alignment}), allowing us to identify the \ac{cad} faces that were masked by the instance segmentation model.

\subsubsection{3D Rendering}
\label{sec:segcad-3drendering}

Many \ac{cad} models are colored in one color, which makes it hard for the human eye and especially harder for a segmentation model to accurately identify the shape and features of the model.
Therefore, given a defined viewing angle, we paint every face of the model visible from the viewing angle randomly in distinct colors and render it via orthographic projection \cite{maynard_drawing_2005}. This ensures that we only see the faces visible from the side we render the model from and not see faces from other sides as it would happen with perspective projection. Lastly, we take a 1920x1080 pixel image of the rendered model. The high resolution of the image ensures capturing small features, like small holes or protrusions, in good detail. Images with a lower resolution make it harder for the segmentation model to detect all features accurately.

\subsubsection{Instance Segmentation}
\label{sec:segcad-inst-seg}


To segment the rendered \ac{cad} model based on a free-text description, we use GroundingDINO \cite{liu_grounding_2024} and \ac{sam} \cite{kirillov_segment_2023}, chosen for their diverse training datasets and strong generalization across domains \cite{noauthor_papers_nodate,noauthor_papers_nodate-1,noauthor_papers_nodate-2}. In contrast, alternatives like Point-LLM \cite{zhou_point-sam_2024} were trained on smaller, specialized datasets, limiting their generalization. Additionally, our approach enables answering view-related questions by rendering the model from specific angles.
GroundingDINO processes the rendered \ac{cad} image and free-text part description, outputting multiple bounding boxes with associated probabilities. We found a 30 \% probability threshold optimal (see Ablation \ref{sec:ablation}). Bounding boxes with probabilities below this threshold are discarded. 
Finally, \ac{sam} \cite{kirillov_segment_2023} refines each bounding box and rendered \ac{cad} image into segmentation masks. This combination of GroundingDINO and \ac{sam} was first proposed by Ren et al. \cite{ren_grounded_2024}.
GroundingDINO tends to select large portions or even the entire \ac{cad} model as a single bounding box. To address this, we filter out masks that cover more than 45 \% of the rendered \ac{cad} model. The segmentation process can be seen in Figure~\ref{fig:architecture}.

\subsubsection{Image to \ac{cad} Model Alignment}
\label{sec:img-to-cad-alignment}

The 2D segmentation masks identify the pixels in the rendered view that correspond to parts matching the part description. To determine the faces of the \ac{cad} model displayed at the masked pixels, we use ray casting \cite{glassner_introduction_1989}. However, because the segmentation mask operates on a 2D rendering of the \ac{cad} model and  has no pixel-level segmentation accuracy, additional post-processing is necessary to refine the selection of faces.

First, segmentation masks often cover small portions of neighboring faces. To filter these out, we only select a face if more than 5 \% of its visible surface in the rendered image is covered by the segmentation mask.

Second, adjacent regions in the rendered \ac{cad} image do not always correspond to adjacent faces on the \ac{cad} model itself, such as when there is content behind a hole. The segmentation mask generated by \ac{sam} sometimes selects faces that are adjacent in the 2D image but not on the actual \ac{cad} model. We require that a segmentation mask includes only faces that are truly adjacent in the 3D space. To enforce this, we perform a post-pruning of the \ac{cad} faces, depicted in Figure \ref{fig:pruning}. Using ray casting, we calculate the distance of each face from the camera's viewport. Starting with the masked face closest to the viewport, we recursively select only those masked faces that are adjacent to it. This ensures that no faces are selected that are further back on the \ac{cad} model and not adjacent to the closer faces. 

\begin{figure}
    \centering
    \includesvg[width=\linewidth]{images/pruning.svg}
    \caption{Pruning \ac{cad} faces to retain only neighboring ones. From raw faces (2) derived from a 2D mask (1), ray casting identifies the closest face (3). Adjacent faces are then selected recursively via the adjacency graph (4), discarding non-adjacent ones, i.e. visible through holes (1). The final pruned part (5) excludes the faces on the right, as they are distant from the viewport (3) and not adjacent to the masked faces on the left.}
    \label{fig:pruning}
\end{figure}

\subsubsection{Multi-View Rendering and View-Specific Retrieval}
\label{sec:segcad-multiview}

We render the \ac{cad} model from multiple views since not all features are visible from a single viewing angle. We found that 6 viewing angles yielded the best results on the \cadds Benchmark (see Ablation \ref{sec:ablation}).

To address view-related questions, such as ``How many protrusions are visible on the right?'', \segcad is designed to retrieve only parts or features visible from specific viewing directions. After identifying all parts or features that globally match the part description, we render the \ac{cad} model from the requested viewing angles (a subset of top, bottom, right, left, front, and back). Using the ray casting method described in \ref{sec:img-to-cad-alignment}, we determine which parts are actually visible from these angles and discard those that are not.

Additionally, due to the orthographic projection, we do not render the \ac{cad} model exactly along the main axes, as many faces would be orthogonal to the viewport and thus not visible in the rendered image. Instead, we slightly perturb the viewing angle by one degree along the azimuth when rendering along the main axes.

\subsection{\querycad: CAD Understanding, Reasoning and Answering}



\querycad is the first system capable of answering questions about the features or parts of \ac{cad} models. A key focus of this system is its ability to provide precise measurements, positions, and other data by enabling \acp{llm} to interact directly with \ac{cad} models. \querycad can accurately retrieve measurements for specific parts of the \ac{cad} model, such as dimensions, radii, center positions, and depths of features. These measurements can be used to filter relevant parts or to form a response.

The system is designed to handle a wide range of free-text questions, whether posed by other deep-learning systems or human engineers, without imposing any constraints on the types of objects or the structure of the questions.

To achieve these properties, we implemented the following approach:
First, the user's query is passed to a code-writing \ac{llm} using a carefully crafted prompt\footnote{See \linktowebsite~for prompts, \cadds benchmark, and examples.}.
The objective of the \ac{llm} is to generate Python code that computes the measurements that answer the question.
For this, the prompt defines Python classes for the \ac{cad} model and the \ac{cad} part. The Python classes have attributes such as extents or the center position, which can be accessed in the code generated by the \ac{llm}.
To enable the \ac{llm} to search in the \ac{cad} model for a specific feature or part, \segcad is integrated via a Python \ac{api}. The \ac{api} calls are parameterized by the \ac{llm} with the free-text part description and potential viewing angles.

The \ac{llm} processes the prompt and generates Python code using \ac{cot} prompting \cite{wei_chain--thought_2022}. This technique enhances model transparency by having the model explain each reasoning step. We observed that \ac{cot} prompting considerably increases the share of correct model responses. The prompt contains 3 high-quality in-context samples.

We observed that constructing a prompt that clearly states the task without any ambiguities is difficult, but highly important to reliably answer user queries with the code written by the \ac{llm}. Especially when reasoning on 3D shapes, these ambiguities are not easy to spot but important to clarify. One example are the \textit{extents} of an object, which can be defined as the full length of the object along global coordinate axes or the distance from the center of the object to its edge. Properties like the \textit{width}, \textit{height}, or \textit{depth} of an object are similarly ambiguous. We curated a prompt to clarify these linguistic ambiguities, which are often featured in user queries. For example, the prompt clarifies to always use full extents unless otherwise stated via \textit{half-extents} and analogously \textit{half-\{width, depth, height\}}.
Moreover, the prompt pushes the \ac{llm} to do any reasoning and conversion of metrics in Python code instead of implicitly converting the metrics, which we found to improve the performance and reduce hallucinations, especially for smaller models like Llama 3.1 8B \cite{dubey_llama_2024}. This is especially important for conversions between units, such as converting meters to millimeters, which the model is prompted to do directly in Python.

The Python code generated by the \ac{llm} is executed, which in turn calls the \segcad model if stated in the code. Examples of \ac{llm} predictions in response to user questions are shown in Figures \ref{fig:page-1} and \ref{fig:architecture}. During execution, the code accesses the properties of the \ac{cad} models' parts (e.g. the radius of a hole or the center of a rod). This enables \querycad to filter and retrieve the relevant properties based on the user query. Finally, the result is returned as a response to the query.

\subsection{\cadds Benchmark}
\begin{figure}
    \centering
    \includesvg[width=\linewidth]{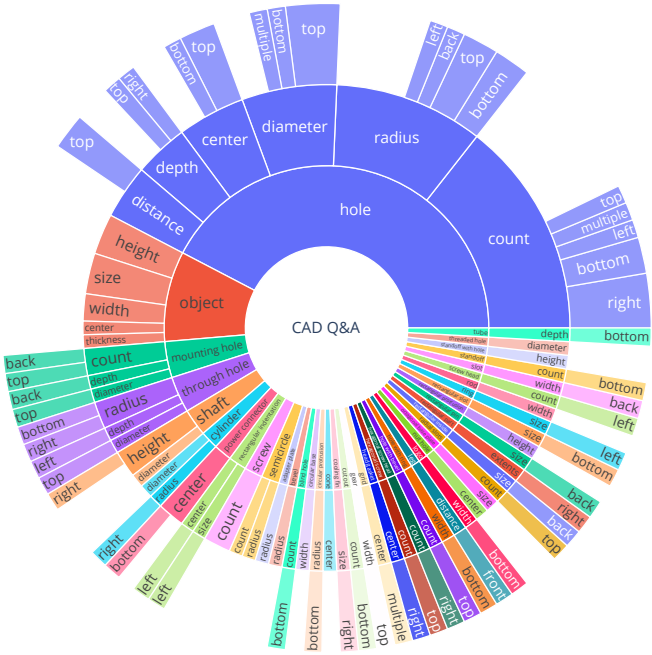}
    \caption{Distribution of dataset questions. Every question is composed of a part it asks about (inner circle), a property to retrieve (middle circle) and optionally one or multiple sides the part must be visible from (outer circle). Some questions enforce additional filtering based on the part properties, like retrieving only parts with a radius of 5 mm.}
    \label{fig:cad-ds-overview}
\end{figure}

At the time of writing, no dataset or benchmark for \ac{cad} question answering exists. To evaluate our approach and establish a common benchmark for future research, we have developed a highly curated and manually annotated dataset specifically tailored for \ac{cad} question answering, available on the paper's website.

Our CAD Q\&A benchmark consists of 18 \ac{cad} models, 10 of which are derived from the ABC dataset \cite{koch_abc_2019}. We specifically selected models from this dataset that are complex and representative of real-world industrial applications. Additionally, we incorporated 8 \ac{cad} models from real-world industrial settings to further enhance the benchmark's relevance.

For each \ac{cad} model, we generated between 4 and 10 questions designed to retrieve specific properties, such as measurements, positions, or counts of particular parts. Initially, we attempted to use \acp{llm} like GPT4o to generate the questions, providing them with screenshots of the \ac{cad} models and a task description. However, the questions generated were often unclear and ambiguous. Therefore, we opted to handcraft the questions for each \ac{cad} model. These questions were then validated by an industrial engineer.

Each question targets either a specific type of part or the entire \ac{cad} model, inquiring about a particular property while sometimes restricting the valid parts by specifying a side the parts should be visible on or applying other filtering criteria based on the part's characteristics. The distribution and structure of these questions are illustrated in Figure~\ref{fig:cad-ds-overview}. To address open-set segmentation, the questions cover a total of 43 diverse parts and ask about 11 different properties.

To define the label for each question, we manually measured the \ac{cad} models using a \ac{cad} kernel and calculated the responses by hand. The final dataset comprises 111 questions across 18 \ac{cad} models.

\section{Results}

\subsection{Evaluation}

We evaluate \querycad on the newly generated \ac{cad} Q\&A Benchmark to quantitatively assess its ability to answer questions about the precise measurements of \ac{cad} models, their parts or features. We use GPT4o \cite{openai_gpt-4_2024} as \ac{llm}, but also compare it to open-source \acp{llm} in Section~\ref{sec:ablation}.

Out of 111 questions, \querycad correctly answered 44, provided partially correct answers where the answer overlaps with the solution for 15 questions, and gave incorrect responses to the remaining 52. These results demonstrate that \querycad can answer a range of questions about \ac{cad} models. To more accurately understand how \querycad fails to answer the 67 wrongly answered samples, we categorize the types of mistakes it made into four groups, detailed in Table \ref{tab:eval-breakdown}.

\textit{Syntax} addresses samples where the \ac{llm} generated invalid Python syntax, which was the case for none of the samples. \textit{Reasoning} corresponds to samples where the proposed Python code features incorrect reasoning or hallucinations, which occurred once. If the Python code is valid, there are still two issues that can cause invalid answers: \textit{Masks}, where incorrect parts were returned by \segcad due to imprecise, missing or incorrect masks proposed by GroundedSAM (see Fig. \ref{fig:viewing-angles} top, left) or due to inaccuracies in the ray casting, which occurred 52 times. Lastly, \textit{\ac{cad}-Interface} classifies answers where the \ac{cad} interface provided in the Python code lacks the necessary capabilities to adequately address the question, which was the case for 14 samples.

\begin{table}
    \centering
    \caption{\querycad on \cadds Benchmark: 111 Questions}
    \label{tab:eval-breakdown}
    \begin{tabular}{l c c c c}
        \toprule
        & Llama 3 & Llama 3.1 & Llama 3.1 & GPT4o \\  
        & 13B & 8B &  405B & Aug-24 \\  \midrule
        Correct             & 32      & 36        & 43 & \textbf{44}       \\
        Wrong & 79      & 75        & 68 & 67       \\
        \hspace{0.75em}Syntax        & 15      & 6         &0 &  0                 \\
        \hspace{0.75em}Reasoning & 17      & 8         & 2 & 1                 \\
        \hspace{0.75em}Masks     & 40      & 51        & 55 & 52                \\
        \hspace{0.75em}CAD-Interface       & 7       & 10        & 11 & 14                \\
        \bottomrule
    \end{tabular}
\end{table}

\subsection{Ablations}
\label{sec:ablation}

\paragraph{LLM Backend} 
To assess the robustness of \querycad across different \acp{llm}, we evaluated it using various models on the \cadds benchmark, as shown in Table \ref{tab:eval-breakdown}. Larger models like GPT4o \cite{openai_gpt-4_2024} and Llama 3.1 405B \cite{dubey_llama_2024} exhibit fewer syntax issues and demonstrate stronger reasoning abilities, leading to 8 more correctly answered questions. There is no significant difference between the predicted answers of Llama 3.1 405B and GPT4o, which shows that \querycad can be used effectively with open source models. Although the large models notably outperform the smaller Llama variants, it is noteworthy that Llama 3.1 8B \cite{dubey_llama_2024}, the smallest available version of Llama 3.1, still shows good reasoning on most samples, resulting in a modest increase of reasoning and syntax errors. This demonstrates that \querycad remains effective even when using lightweight and ultra-fast \acp{llm} like Llama 3.1 8B, making it viable for deployment on edge devices.

\begin{figure}
    \centering
    \includesvg[width=\linewidth]{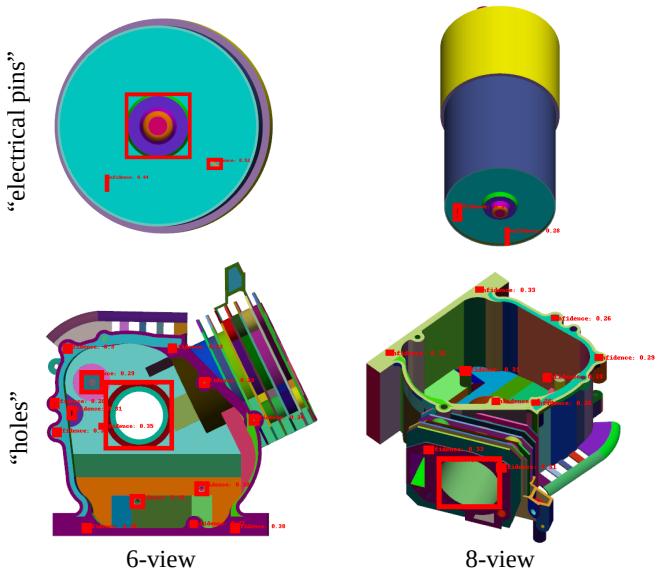}
    \caption{We evaluate \segcad by rendering the \ac{cad} model from either 6 views (left) or 8 views (right). GroundingDINO detects more parts when rendered along the main axes (lower left), but it occasionally selects parts that do not match the description (upper left). In contrast, rendering the model from 8 viewing angles often results in some parts not being selected (lower right) but reduces the likelihood of selecting incorrect parts (upper right).}
    \label{fig:viewing-angles}
\end{figure}

\paragraph{Viewing Angles} 
We evaluated \segcad with two different viewing angle configurations: 6 viewing angles along the main axes and 8 viewing angles from all 8 corners of the \ac{cad} model with an azimuth and elevation of the camera of 45 degrees. Figure \ref{fig:viewing-angles} compares the two viewing configurations.
While the configuration with 6 viewing angles makes it harder for GroundedSAM \cite{ren_grounded_2024} to understand the depth of the \ac{cad} model, as most of the features are orthogonal to the viewport, it effectively allows detecting features like through-holes or other geometrical features specifically from that axis. The configuration with 8 viewing angles has the benefit to capture the 3D shape of the model more accurately, reducing the likelihood of wrong predictions by GroundedSAM. One example are shafts and holes, which look the same from the top. However, we observed that with 8 viewing angles, GroundedSAM occasionally misses certain parts, as shown in Figure~\ref{fig:viewing-angles}. Additionally, it becomes more challenging to select the parts visible from a specific viewing direction with 8 viewing angles.

On the \cadds dataset, using 6 viewing angles resulted in 44 correctly answered questions, while \querycad answered 28 answers correctly with 8 viewing angles. This mainly stems from the fact that \segcad with 8 viewing angles often did not detect all parts the questions asked about or returned parts that were not visible from the sides considered in the question.

\paragraph{GroundingDINO Threshold} 

In \segcad, the likelihood threshold for GroundingDINO's bounding box suggestions is a critical hyperparameter. Bounding boxes with a likelihood below this threshold are discarded. The threshold manages the trade-off between detecting parts with low scores, such as those that are only partially visible or very small, and the risk of selecting incorrect parts when the threshold is too low. We found that a threshold of 30 \% produced the best results on the \cadds benchmark with 44 correct answers compared to 41 correct answers with a threshold of 25 \%.


\subsection{Validation}

\querycad can be integrated in existing deep-learning based methods in robotics to retrieve information directly from \ac{cad} models given a free-form question. We validate this by combining \querycad with the MetaWizard program synthesis framework \cite{alt_robogrind_2024}. During program synthesis, MetaWizard asks the user to define parameters of the robot program that cannot be derived automatically from its internal knowledge base. The integration of \querycad permits MetaWizard to directly ask \querycad, instead of the human user, in natural language about the parameters it needs.

MetaWizard with \querycad is employed to program an assembly task where the robot picks up a gear and inserts it into an engine block, as shown in Figure \ref{fig:page-1}. The user can program the robot for this task in natural language. During the program synthesis, MetaWizard queries \querycad twice: first, to determine the center of the gear by asking ``What is the center of the object?'' on the gear's \ac{cad} model, and second, to locate the tip of the shaft where the gear must be inserted by asking ``Where is the tip of the shaft?'' on the engine block's \ac{cad} model. For details of the interaction and results, we refer to the paper's website.

\section{Conclusion and Outlook}
\label{sec:conclusion}

We introduce \querycad, the first system for question answering tasks on \ac{cad} models. Our method accurately identifies the parts of the \ac{cad} model referenced in a question and retrieves precise measurements by grounding its predictions in a \ac{cad} model. We demonstrate the effectiveness of our approach in handling open-vocabulary queries related to \ac{cad} models. To assess its performance, we develop and publish the first \cadds benchmark, a tool that future research can utilize for comparative evaluation. Furthermore, by integrating our system into an existing robot program synthesis method, we validate that it successfully provides automatic querying of \ac{cad} information to existing algorithms.

While \querycad performs well on the \cadds benchmark, there is still room for improvement. The robustness of our approach largely depends on the accuracy of part segmentation of the \ac{cad} model with \segcad. The most common errors during question answering arise from incorrect or missed part detection and artifacts caused by ray casting. Enhancing the instance segmentation of \segcad remains an open area for future research. Additionally, while \querycad can handle a wide range of queries, such as measurements, e.g. radius, diameter, width, and part counts, it currently calculates all metrics in world coordinates. This can lead to incorrect answers when questions pertain to a part's local orientation. Moreover, properties like surface normals are not yet supported, presenting another avenue for future research.

\bibliographystyle{IEEEtran}
\bibliography{bibliography_config,bibliography}

\end{document}